\documentclass[11pt]{article}
\usepackage{konvens2019}
\usepackage{mathptmx}
\usepackage[scaled=.90]{helvet}
\usepackage{courier}

\makeatletter
\newcommand{\@BIBLABEL}{\@emptybiblabel}
\newcommand{\@emptybiblabel}[1]{}
\makeatother
\usepackage{hyperref}
\usepackage{amsmath,amssymb,amsfonts}
\usepackage{graphicx}
\usepackage{xcolor}
\usepackage{tabularx}
\usepackage{xspace}
\usepackage{booktabs}
\usepackage{url}
\newcommand{\F}{$\text{F}_1$\xspace}
\newcommand{\ccol}[1]{\multicolumn{1}{c}{#1}}
\newcommand{\ccolr}[1]{\multicolumn{1}{c}{\rotatebox{90}{#1}}}

\newcommand{\eg}{\textit{e.\,g.}\xspace}
\usepackage{url}
\usepackage{latexsym}
\usepackage{scalefnt}

\title{Towards Multimodal Emotion Recognition in\\ German Speech Events in Cars
  using Transfer Learning}

\author{Deniz Cevher$^{1,2}$\thanks{ \ \ The first two authors contributed equally.}, Sebastian Zepf$^{1*}$ \and Roman Klinger$^2$ \\
	$^1$ Mercedes-Benz Research \& Development, Daimler AG, Sindelfingen, Germany\\
	$^2$ Institut f\"ur Maschinelle Sprachverarbeitung, University of Stuttgart, Germany \\
  {\tt \{firstname.lastname\}@daimler.com}\\
  {\tt \{firstname.lastname\}@ims.uni-stuttgart.de}\\
}

\date{}

\begin{document}
\maketitle
\begin{abstract}
  The recognition of emotions by humans is a complex process which
  considers multiple interacting signals such as facial expressions
  and both prosody and semantic content of utterances. Commonly,
  research on automatic recognition of emotions is, with few
  exceptions, limited to one modality. We describe an in-car
  experiment for emotion recognition from speech interactions for
  three modalities: the audio signal of a spoken interaction, the
  visual signal of the driver's face, and the manually transcribed
  content of utterances of the driver. We use off-the-shelf tools for
  emotion detection in audio and face and compare that to a neural
  transfer learning approach for emotion recognition from text which
  utilizes existing resources from other domains. We see that transfer
  learning enables models based on out-of-domain corpora to perform
  well. This method contributes up to 10 percentage points in \F, with
  up to 76 micro-average \F across the emotions joy, annoyance and
  insecurity. Our findings also indicate that off-the-shelf-tools
  analyzing face and audio are not ready yet for emotion detection in
  in-car speech interactions without further adjustments.
\end{abstract}


\section{Introduction}
Automatic emotion recognition is commonly understood as the task of
assigning an emotion to a predefined instance, for example an
utterance (as audio signal), an image (for instance with a depicted
face), or a textual unit (e.g., a transcribed utterance, a sentence,
or a Tweet). The set of emotions is often following the original
definition by Ekman~\shortcite{Ekman1992}, which includes anger, fear,
disgust, sadness, joy, and surprise, or the extension by Plutchik
\shortcite{Plutchik1980} who adds trust and anticipation.

Most work in emotion detection is limited to one modality. Exceptions
include \newcite{Busso2004} and \newcite{Sebe2005}, who investigate
multimodal approaches combining speech with facial
information. Emotion recognition in speech can utilize semantic
features as well \cite{Anagnostopoulos2015}. Note that the term
``multimodal'' is also used beyond the combination of vision, audio,
and text. For example, \newcite{Soleymani2012} use it to refer to the
combination of electroencephalogram, pupillary response and gaze
distance.

In this paper, we deal with the specific situation of car environments
as a testbed for multimodal emotion recognition. This is an
interesting environment since it is, to some degree, a controlled
environment: Dialogue partners are limited in movement, the degrees of
freedom for occurring events are limited, and several sensors which
are useful for emotion recognition are already integrated in this setting.
More specifically, we focus on emotion
recognition from speech events in a dialogue with a human partner and
with an intelligent agent.

Also from the application point of view, the domain is a relevant
choice: Past research has shown that emotional intelligence is
beneficial for human computer interaction. Properly processing
emotions in interactions increases the engagement of users and can
improve performance when a specific task is to be fulfilled
\cite{Klein2002,Coplan2011,Partala2004,Pantic2005}. This is mostly
based on the aspect that machines communicating with humans appear to
be more trustworthy when they show empathy and are perceived as being
natural \cite{Partala2004,Brave2005,Pantic2005}.

Virtual agents play an increasingly important role in the automotive
context and the speech modality is increasingly being used in cars due
to its potential to limit distraction. It has been shown that adapting
the in-car speech interaction system according to the drivers'
emotional state can help to enhance security, performance as well as
the overall driving experience
\cite{Nass2005,Harris2010}.

With this paper, we investigate how each of the three considered
modalitites, namely facial expressions, utterances of a driver as an
audio signal, and transcribed text contributes to the task of emotion
recognition in in-car speech interactions. We focus on the five
emotions of \textit{joy}, \textit{insecurity}, \textit{annoyance},
\textit{relaxation}, and \textit{boredom} since terms corresponding to
so-called fundamental emotions like \textit{fear} have been shown to
be associated to too strong emotional states than being appropriate
for the in-car context \cite{Dittrich2019}. Our first contribution is
the description of the experimental setup for our data
collection. Aiming to provoke specific emotions with situations which
can occur in real-world driving scenarios and to induce speech
interactions, the study was conducted in a driving simulator. Based on
the collected data, we provide baseline predictions with off-the-shelf
tools for face and speech emotion recognition and compare them to a
neural network-based approach for emotion recognition from text. Our
second contribution is the introduction of transfer learning to adapt
models trained on established out-of-domain corpora to our use
case. We work on German language, therefore the transfer consists of a
domain and a language transfer.

\section{Related Work}
\subsection{Facial Expressions}
A common approach to encode emotions for facial expressions is the
facial action coding system FACS
\cite{Ekman1978,gunawan2015face,lien1998automated}. As the reliability
and reproducability of findings with this method have been critically
discussed \cite{Mesman2012}, the trend has increasingly shifted to
perform the recognition directly on images and videos, especially with
deep learning. For instance, \newcite{jung2015joint} developed a model
which considers temporal geometry features and temporal appearance
features from image sequences. \newcite{kim2016hierarchical} propose
an ensemble of convolutional neural networks which outperforms
isolated networks.

In the automotive domain, FACS is still popular. \newcite{Ma2017} use
support vector machines to distinguish \textit{happy},
\textit{bothered}, \textit{confused}, and \textit{concentrated} based
on data from a natural driving environment. They found that
\textit{bothered} and \textit{confused} are difficult to distinguish,
while \textit{happy} and \textit{concentrated} are well
identified. Aiming to reduce computational cost, \newcite{Tews2011}
apply a simple feature extraction using four dots in the face defining
three facial areas. They analyze the variance of the three facial
areas for the recognition of \textit{happy}, \textit{anger} and
\textit{neutral}. \newcite{Ihme2018} aim at detecting
\textit{frustration} in a simulator environment. They induce the
emotion with specific scenarios and a demanding secondary task and are
able to associate specific face movements according to FACS.
\newcite{Paschero2012} use OpenCV (\url{https://opencv.org/}) to
detect the eyes and the mouth region and track facial movements. They
simulate different lightning conditions and apply a multilayer
perceptron for the classification task of Ekman's set of fundamental
emotions.

Overall, we found that studies using facial features usually focus on
continuous driver monitoring, often in driver-only scenarios. In
contrast, our work investigates the potential of emotion
recognition during speech interactions.

\subsection{Acoustic}
Past research on emotion recognition from acoustics mainly
concentrates on either feature selection or the
development of appropriate classifiers. \newcite{rao2013emotion} as
well as \newcite{ververidis2004automatic} compare local and global
features in support vector machines. Next to such discriminative
approaches, hidden Markov models are well-studied, however, there is
no agreement on which feature-based classifier is most suitable
\cite{el2011survey}. Similar to the facial expression modality, recent
efforts on applying deep learning have been increased for acoustic
speech processing. For instance, \newcite{lee2015high} use a recurrent
neural network and \newcite{palaz2015analysis} apply a convolutional
neural network to the raw speech signal. \newcite{Neumann2017} as well
as \newcite{Trigeorgis2016} analyze the importance of features in the
context of deep learning-based emotion recognition.

In the automotive sector, \newcite{Boril2011} approach the detection
of negative emotional states within interactions between driver and
co-driver as well as in calls of the driver towards the automated
spoken dialogue system. Using real-world driving data, they find that
the combination of acoustic features and their respective Gaussian
mixture model scores performs best. \newcite{Schuller2006} collects
2,000 dialog turns directed towards an automotive user interface and
investigate the classification of \textit{anger}, \textit{confusion},
and \textit{neutral}. They show that automatic feature generation and
feature selection boost the performance of an SVM-based
classifier. Further, they analyze the performance under systematically
added noise and develop methods to mitigate negative effects.
For more details, we refer the reader to the survey by
\newcite{Schuller2018}. In this work, we explore the straight-forward
application of domain independent software to an in-car scenario
without domain-specific adaptations.

\subsection{Text}
Previous work on emotion analysis in natural language processing
focuses either on resource creation or on emotion classification for a
specific task and domain. On the side of resource creation, the early
and influential work of \newcite{Pennebaker2015} is a dictionary of
words being associated with different psychologically relevant
categories, including a subset of emotions. Another popular resource
is the NRC dictionary by \newcite{Mohammad2012b}. It contains more
than 10000 words for a set of discrete emotion classes. Other
resources include WordNet Affect~\cite{Strapparava2004} which
distinguishes particular word classes.  Further, annotated corpora
have been created for a set of different domains, for instance fairy
tales \cite{Ovesdotter2005}, Blogs \cite{Aman2007}, Twitter
\cite{Mohammad2017c,Schuff2017,Mohammad2012,Mohammad2017,Klinger2018x},
Facebook \cite{Preotiuc2016}, news headlines \cite{Strapparava2007},
dialogues \cite{Li2017}, literature \cite{Kim2017a}, or self reports
on emotion events \cite{Scherer1997} (see \cite{Bostan2018} for an
overview).

To automatically assign emotions to textual units, the application of
dictionaries has been a popular approach and still is, particularly in
domains without annotated corpora. Another approach to overcome the
lack of huge amounts of annotated training data in a particular domain
or for a specific topic is to exploit distant supervision: use the
signal of occurrences of emoticons or specific hashtags or words to
automatically label the data. This is sometimes referred to as
self-labeling \cite{Klinger2018x,Pool2016,Felbo2017,Wang2012b}.

A variety of classification approaches have been tested, including
SNoW \cite{Ovesdotter2005}, support vector machines \cite{Aman2007},
maximum entropy classification, long short-term memory network, and
convolutional neural network models
\cite[\textit{i.a.}]{Schuff2017}. More recently, the state of the art
is the use of transfer learning from noisy annotations to more
specific predictions \cite{Felbo2017}. Still, it has been shown that
transferring from one domain to another is challenging, as the way
emotions are expressed varies between areas \cite{Bostan2018}. The
approach by \newcite{Felbo2017} is different to our work as they use a
huge noisy data set for pretraining the model while we use small high
quality data sets instead.

Recently, the state of the art has also been pushed forward with a set
of shared tasks, in which the participants with top results mostly
exploit deep learning methods for prediction based on pretrained
structures like embeddings or language models
\cite{Klinger2018x,Mohammad2018,Mohammad2017}.

Our work follows this approach and builds up on embeddings with deep
learning.  Furthermore, we approach the application and adaption of
text-based classifiers to the automotive domain with transfer
learning.

\begin{figure}
  \centering
  \includegraphics[width=0.9\linewidth]{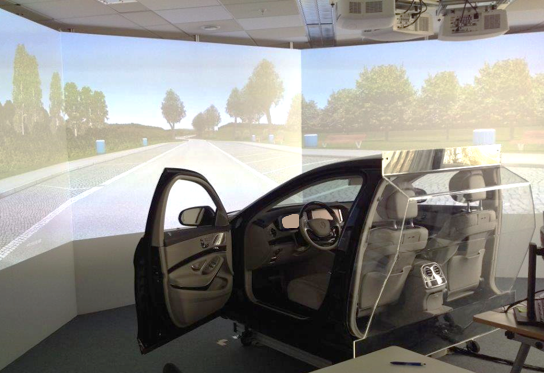}
  \caption{The setup of the driving simulator.}
  \label{fig:dataset:method:simulator}
\end{figure}

\section{Data set Collection}
\label{sec:dataset}
The first contribution of this paper is the construction of the AMMER data set 
which we describe in the following. We focus
on the drivers' interactions with both a virtual agent as well as a
co-driver. To collect the data in a safe and controlled environment
and to be able to consider a variety of predefined driving situations,
the study was conducted in a driving simulator. 

\subsection{Study Setup and Design}
\label{sec:dataset:setup}
The study environment consists of a fixed-base driving simulator
running Vires's VTD (Virtual Test Drive, v2.2.0) simulation software
(\url{https://vires.com/vtd-vires-virtual-test-drive/}). The vehicle
has an automatic transmission, a steering wheel and gas and brake
pedals. We collect data from video, speech and biosignals (Empatica E4
to record heart rate, electrodermal activity, skin temperature, not
further used in this paper) and questionnaires. Two RGB cameras are
fixed in the vehicle to capture the drivers face, one at the sun
shield above the drivers seat and one in the middle of the
dashboard. A microphone is placed on the center console. One
experimenter sits next to the driver, the other behind the
simulator. The virtual agent accompanying the drive is realized as
Wizard-of-Oz prototype which enables the experimenter to manually
trigger prerecorded voice samples playing trough the in-car speakers
and to bring new content to the center screen. Figure
\ref{fig:dataset:method:simulator} shows the driving simulator.

\begin{table}[t]
  \centering
  \setlength{\tabcolsep}{3pt}
  \begin{tabularx}{\linewidth}{p{14.55mm}X}
    \toprule
    Type & Example \\
    \cmidrule(r){1-1}\cmidrule(l){2-2}
    D--A, beginning  & Wie geht es dir gerade und wie sind deine Gedanken zur bevorstehenden Fahrt? 
                     \textit {How are you doing right now? What are
                       your thoughts about the upcoming drive?} \\
    D--A, reaching destination & Bei \"uber 50 Teilnehmern hast du
    die zweitschnellste Zeit erreicht. Was glaubst du? Wie hast du es
    geschafft so schnell zu sein?
    \textit{ Among more than 50 participants you achieved the second
    best result. What do you think? How did you manage to achieve
    that?}\\
  D--A, after driving & Du hast im letzten Streckenabschnitt ein paar Mal
  stark gebremst. Was ist da passiert?
  \textit{ In the last section, you slowed down multiple
    times. What happened?}\\
  D--Co, low-demand section & Erinnern Sie sich an Ihren letzten Urlaub. Bitte
  beschreiben Sie, wie dieser Urlaub f\"ur Sie war?
  \textit{ Remember your last vacation. Please describe how it was.}\\
    \bottomrule    
  \end{tabularx}
  \caption{Examples for triggered interactions with translations to English. (D: Driver, A: Agent,
    Co: Co-Driver)}
  \label{tab:interactionexamples}
\end{table}

The experimental setting is comparable to an everyday driving
task. Participants are told that the goal of the study is to evaluate
and to improve an intelligent driving assistant. To increase the
probability of emotions to arise, participants are instructed to reach
the destination of the route as fast as possible while following
traffic rules and speed limits. They are informed that the time needed
for the task would be compared to other participants. The route
comprises highways, rural roads, and city streets. A navigation system
with voice commands and information on the screen keeps the
participants on the predefined track.

To trigger emotion changes in the participant, we use the following
events: (i) a car on the right lane cutting off to the left lane when
participants try to overtake followed by trucks blocking both lanes
with a slow overtaking maneuver (ii) a skateboarder who appears
unexpectedly on the street and (iii) participants are praised for
reaching the destination unexpectedly quickly in comparison to
previous participants.

Based on these events, we trigger three interactions
(Table~\ref{tab:interactionexamples} provides examples) with the
intelligent agent (\textit{Driver-Agent Interactions,
  D--A}). Pretending to be aware of the current situation, \eg, to
recognize unusual driving behavior such as strong braking, the agent
asks the driver to explain his subjective perception of these events
in detail.  Additionally, we trigger two more interactions with the
intelligent agent at the beginning and at the end of the drive, where
participants are asked to describe their mood and thoughts regarding
the (upcoming) drive. This results in five interactions between the
driver and the virtual agent.

Furthermore, the co-driver asks three different questions during
sessions with light traffic and low cognitive demand
(\textit{Driver-Co-Driver Interactions, D--Co}). These questions are
more general and non-traffic-related and aim at triggering the
participants' memory and fantasy. Participants are asked to describe
their last vacation, their dream house and their idea of the perfect
job. In sum, there are eight interactions per participant (5 D--A, 3
D--Co).

\begin{table*}[t]
  \centering\small\scalefont{1.05}
  \begin{tabularx}{\linewidth}{lcX}
    \toprule
    E & IT & Example \\
    \cmidrule(r){1-1}\cmidrule(r){2-2}\cmidrule(r){3-3}
    J  & A  & Ich glaube, weil ich ziemlich schnell auf Situationen
    reagieren kann, weil ich eine ziemlich gute Reaktion habe. Und ich
    w\"urde auch behaupten, dass ich relativ vorausschauend fahre,
    weil ich schon einiges an Fahrerfahrung mitbringe. 
    \textit{I think because I can respond to situations very quickly
      because my reaction is very good. And I would say that I drive
      foresightful because I have a lot of driving experience.} \\
    J  & C  & Letzter Urlaub war im September 2018. Singapur und
    Bali. War sehr sch\"on. Erholung, andere Kultur, andere
    L\"ander. War sehr gut und ist zu wiederholen. 
    \textit{Last vacation was in September 2018. Singapore and
      Bali. It was beautiful. Recreation, different culture, different
      countries. It was very good and needs repetition.} \\
    A  & A  & Zwei bis drei Mal Fahrzeuge, die Kolonne fuhren. Und
    das letzte Fahrzeug hat, f\"ur mein Gef\"uhl, sehr ruckartig und
    mit wenig nach hinten zu schauen, die Spur gewechselt und mich
    dazu gezwungen, dann doch noch meine Geschwindigkeit zu
    reduzieren. 
    \textit{Two or three times vehicles were driving behind
      each other. The last vehicle cut off my lane, in my opinion
      very quickly and without looking back and forced me to slow
      down.} \\
    A  & C  & Mir geht es nicht besonders gut. Die Fahrt war sehr
    stressig. Ich schwitze ziemlich. 
    \textit{I'm not feeling well. The ride was stressful. I am
      sweating.}\\
    I  & A  & Letzter Urlaub war nicht so gut f\"ur
    mich. Obwohl. Naja doch. Der letzte war schon wieder gut. Das war
    im Sommer. Da war es n\"amlich so abartig warm dieses Jahr. Und
    wir haben bei uns daheim. Also ich komme ja vom Land. Wir haben
    bei uns daheim auf dem Land unseren Wohnwagen ausgebaut. 
    \textit{Last vacation was not so good for me. Although. Well,
  yes. The last one was good. It was in summer. It was very warm this
  year. And we have at home. I come from the countryside. We have
  furnished our mobile home.} \\
    I  & C  & Ein Mensch ist \"uber die Stra{\ss}e gelaufen und ich
    habe ihn zuerst nicht gesehen. 
    \textit{A human crossed the street and I haven't seen him in the
  first moment.} \\
    B  & A  & Ich habe mich immer an die Richtgeschwindigkeit
    gehalten. Und ja. Ich wei{\ss} auch nicht. 
    \textit{I always followed the recommended velocity. And, well. I don't know.} \\
    B  & C  & Ja. Nicht viel arbeiten und viel Geld verdienen. 
    \textit{Yes. Not working much and earning a lot of money.} \\
    R  & A  & Mir geht es gut und ich bin gespannt auf die
    Fahrt. Ich denke, es macht Spa{\ss}. 
    \textit{I am fine and I am looking forward to the ride. I think
  it will be fun.} \\
    R  & C  & Ja, ich erinnere mich an den letzten Urlaub und der
    war sch\"on, war erholsam und war warm. 
    \textit{Yes, I remember the last vacation. It was nice,
  recreative and warm.} \\
    N  & A  & Es sind Autos von der rechten Spur auf meine Spur
    gezogen, welche davor deutlich langsamer waren. 
    \textit{Cars were changing into my lane, which were slower
  before.} \\
    N  & C  & Ein Haus, das relativ alleine f\"ur sich steht. Am
    besten am Meer und mit einem gr\"unen Garten. Und ja. Viel Platz
    f\"ur sich. 
    \textit{A house with space around. In the best case at the
  sea and with a green garden. And yes. A lot of space for us.}
\\
    \bottomrule
  \end{tabularx}
  \caption{Examples from the collected data set (with translation to
    English). E: Emotion, IT: interaction type with agent (A)
    and with Codriver (C). J: Joy, A: Annoyance, I: Insecurity, B:
    Boredom, R: Relaxation, N: No emotion.}
  \label{tab:instanceexamples}
\end{table*}

\subsection{Procedure}
\label{sec:dataset:procedure}

At the beginning of the study, participants were welcomed and the
upcoming study procedure was explained. Subsequently, participants
signed a consent form and completed a questionnaire to provide
demographic information. After that, the co-driving experimenter
started with the instruction in the simulator which was followed by a
familiarization drive consisting of highway and city driving and
covering different driving maneuvers such as tight corners, lane
changing and strong braking. Subsequently, participants started with
the main driving task. The drive had a duration of 20 minutes
containing the eight previously mentioned speech interactions. After
the completion of the drive, the actual goal of improving automatic
emotional recognition was revealed and a standard emotional
intelligence questionnaire, namely the
TEIQue-SF~\cite{cooper2010psychometric}, was handed to the
participants. Finally, a retrospective interview was conducted, in
which participants were played recordings of their in-car interactions
and asked to give discrete (annoyance, insecurity, joy, relaxation,
boredom, none, following~\cite{Dittrich2019}) was well as dimensional
(valence, arousal, dominance~\cite{Posner2005} on a 11-point scale)
emotion ratings for the interactions and the according situations. We
only use the discrete class annotations in this paper.

\subsection{Data Analysis}
Overall, 36 participants aged 18 to 64 years ($\mu$=28.89,
$\sigma$=12.58) completed the experiment. This leads to 288
interactions, 180 between driver and the agent and 108 between driver
and co-driver.  The emotion self-ratings from the participants yielded
90 utterances labeled with \textit{joy}, 26 with \textit{annoyance},
49 with \textit{insecurity}, 9 with \textit{boredom}, 111 with
\textit{relaxation} and 3 with \textit{no emotion}. One example
interaction per interaction type and emotion is shown in
Table~\ref{tab:instanceexamples}. For further experiments, we only use
joy, annoyance/anger, and insecurity/fear due to the small sample size
for boredom and no emotion and under the assumption that relaxation
brings little expressivity.

\section{Methods}
\subsection{Emotion Recognition from Facial Expressions}
We preprocess the visual data by extracting the sequence of images for
each interaction from the point where the agent's or the co-driver's
question was completely uttered until the driver's response
stops. The average length is 16.3 seconds, with the minimum at 2.2s
and the maximum at 54.7s. We apply an off-the-shelf tool for emotion
recognition (the manufacturer cannot be disclosed due to licensing
restrictions). It delivers frame-by-frame scores ($\in [0;100]$) for
discrete emotional states of \textit{joy}, \textit{anger} and
\textit{fear}. While \textit{joy} corresponds directly to our
annotation, we map \textit{anger} to our label \textit{annoyance} and
\textit{fear} to our label \textit{insecurity}. The maximal average
score across all frames constitutes the overall classification for the video
sequence. Frames where the software is not able to detect the face are
ignored.

\subsection{Emotion Recognition from Audio Signal}
We extract the audio signal for the same sequence as described for
facial expressions and apply an off-the-shelf tool for emotion
recognition. The software delivers single classification scores for a
set of 24 discrete emotions for the entire utterance. We consider the
outputs for the states of joy, anger, and fear, mapping analogously to
our classes as for facial expressions. Low-confidence predictions are
interpreted as ``no emotion''. We accept the emotion with the highest
score as the discrete prediction otherwise.

\begin{table*}
\centering
\setlength{\tabcolsep}{10pt}
\begin{tabular}{lrrrrr}
\toprule
Data set & Fear & Anger& Joy & Total\\
\cmidrule(r){1-1}\cmidrule(rl){2-2}\cmidrule(rl){3-3}\cmidrule(rl){4-4}\cmidrule(l){5-5}
Figure8
& 8,419 & 1,419 & 9,179 & 19,017\\
EmoInt
& 2,252 & 1,701 & 1,616 & 5,569\\
ISEAR
& 1,095 & 1,096 & 1,094 & 3,285\\
TEC
& 2,782 & 1,534 & 8,132 & 12,448\\
AMMER
& 49 & 26 & 90 & 165 \\
\bottomrule
\end{tabular}
\caption{Class distribution of the used data sets for the considered
  emotional states (Figure8 \protect\cite{CFDataSet2016}, EmoInt
  \protect\cite{mohammad2017a}, ISEAR, \protect\cite{Scherer1997}, TEC
  \protect\cite{Mohammad2012}, AMMER (this paper)).}
\label{tab:datasets}
\end{table*}

\begin{figure}
  \centering
  \includegraphics[width=0.95\linewidth]{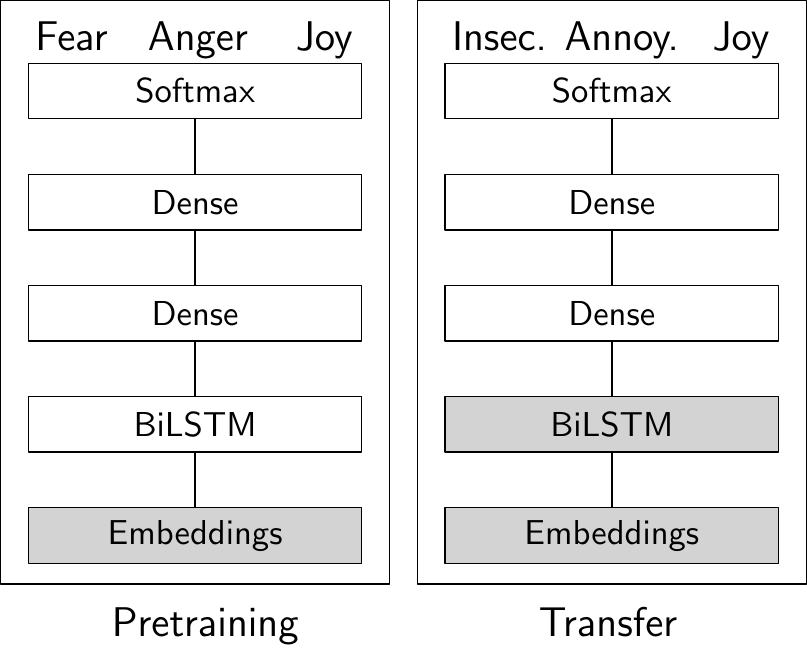}
  \caption{Model for Transfer Learning from Text. Grey boxes contain
    frozen parameters in the corresponding learning step.}
  \label{fig:transfer}
\end{figure}

\subsection{Emotion Recognition from Transcribed Utterances}
\label{sec:methodstext}
For the emotion recognition from text, we manually transcribe all utterances 
of our AMMER study.
To exploit existing and available data sets which are larger than the
AMMER data set, we develop a transfer learning
approach. We use a neural network with an embedding layer (frozen
weights, pre-trained on Common Crawl and
Wikipedia~\cite{grave2018learning}), a bidirectional
LSTM~\cite{Schuster1997}, and two dense layers followed by a soft max
output layer. This setup is inspired by~\cite{andryushechkin2017}. We
use a dropout rate of 0.3 in all layers and optimize with Adam
\cite{Kingma2015AdamAM} with a learning rate of $10^{-5}$ (These parameters are the same for all further experiments). We build on
top of the Keras library with the TensorFlow backend.  We consider
this setup our \textit{baseline model}.

We train models on a variety of corpora, namely the common format
published by~\cite{Bostan2018} of the \textit{FigureEight} (formally
known as Crowdflower) data set of
social media, the \textit{ISEAR} data \cite{scherer1994}
(self-reported emotional events), and, the Twitter Emotion Corpus
(TEC, weakly annotated Tweets with \#anger, \#disgust, \#fear,
\#happy, \#sadness, and \#surprise, \newcite{Mohammad2012}). From all
corpora, we use instances with labels fear, anger, or joy.  These
corpora are English, however, we do predictions on German
utterances. Therefore, each corpus is preprocessed to German with
Google Translate\footnote{\url{http://translate.google.com}, performed
  on January 4, 2019}. We remove URLs, user tags (``@Username''),
punctuation and hash signs.  The distributions of the data sets are
shown in Table~\ref{tab:datasets}.

To adapt models trained on these data, we apply \textit{transfer
  learning} as follows: The model is first trained until convergence
on one out-of-domain corpus (only on classes fear, joy, anger for
compatibility reasons). Then, the parameters of the bi-LSTM layer are
frozen and the remaining layers are further trained on AMMER. This
procedure is illustrated in Figure~\ref{fig:transfer}

\begin{table}[t]
  \centering
  \setlength{\tabcolsep}{8pt}
  \begin{tabular}{lrrrrr}
    \toprule
    & \multicolumn{4}{c}{Vision} & \\
    \cmidrule(lr){2-5}
    & \ccolr{Fear} & \ccolr{Anger} & \ccolr{Joy} &  & \ccolr{Total} \\
    \cmidrule(rl){2-2}\cmidrule(rl){3-3}\cmidrule(lr){4-4}\cmidrule(rl){5-5}\cmidrule(l){6-6}
    Insecurity & 11 & 17 & 21 && 49\\
    Annoyance  & 10 & 7  & 9  && 26\\
    Joy        & 24 & 27 & 39 && 90\\
    Total      & 45 & 51 & 69 && 165\\[1mm]
    \cmidrule(rl){1-6}
    & \multicolumn{4}{c}{Audio} & \\
    \cmidrule(lr){2-5}
    & \ccolr{Fear} & \ccolr{Anger} & \ccolr{Joy} & \ccolr{No} & \ccolr{Total} \\
    \cmidrule(rl){2-2}\cmidrule(rl){3-3}\cmidrule(lr){4-4}\cmidrule(rl){5-5}\cmidrule(l){6-6}
    Insecurity & 17 & 14 & 1 & 17 & 49\\
    Annoyance  & 12 & 7  & 0 & 7 & 26\\
    Joy        & 27 & 26 & 4 & 33 & 90\\
    Total      & 56 & 47 & 5 & 57 & 165 \\ 
    \cmidrule(rl){1-6}
    & \multicolumn{4}{c}{Transfer Learning Text} & \\
    \cmidrule(lr){2-5}
    & \ccolr{Fear} & \ccolr{Anger} & \ccolr{Joy} & \ccolr{No} & \ccolr{Total} \\
    \cmidrule(rl){2-2}\cmidrule(rl){3-3}\cmidrule(lr){4-4}\cmidrule(rl){5-5}\cmidrule(l){6-6}
    Insecurity & 33 & 0 & 16 && 49 \\
    Annoyance  & 7 & 4 & 15 && 26\\
    Joy        & 1 & 1 & 88 && 90\\
    Total      & 41 & 5 & 119 && 165\\
    \bottomrule
  \end{tabular}
  \caption{Confusion Matrix for Face Classification and
    Audio Classification (on full AMMER data) and for transfer
    learning from text (training set of EmoInt and test set of
    AMMER). Insecurity, annoyance and joy are the gold labels. Fear, anger and joy are predictions.}
  \label{tab:confusion}
\end{table}

\begin{table*}
  \centering
  \setlength{\tabcolsep}{14pt}
  \begin{tabular}{l rrr rrr rrr}
    \toprule
    & \multicolumn{3}{c}{Vision} & \multicolumn{3}{c}{Audio} & \multicolumn{3}{c}{Text (TL)} \\
    \cmidrule(r){2-4}\cmidrule(rl){5-7}\cmidrule(l){8-10}
    & \ccol{P} & \ccol{R} & \ccol{\F}& \ccol{P} & \ccol{R} & \ccol{\F}& \ccol{P} & \ccol{R} & \ccol{\F}\\
    \cmidrule(rl){2-2}\cmidrule(rl){3-3}\cmidrule(lr){4-4}\cmidrule(rl){5-5}\cmidrule(rl){6-6}\cmidrule(lr){7-7}\cmidrule(rl){8-8}\cmidrule(rl){9-9}\cmidrule(lr){10-10}
    Insecurity & 24 & 22 & 23 & 31 & 35 & 33 & 80 & 67 & 73 \\
    Annoyance  & 14 & 39 & 21 & 15 & 27 & 19 &  80 & 15  &  26 \\
    Joy        & 57 & 43 & 49 & 80 &  4 &  8 & 74 & 98 & 84 \\
\cmidrule(r){1-1}\cmidrule(rl){2-2}\cmidrule(rl){3-3}\cmidrule(lr){4-4}\cmidrule(rl){5-5}\cmidrule(rl){6-6}\cmidrule(lr){7-7}\cmidrule(rl){8-8}\cmidrule(rl){9-9}\cmidrule(lr){10-10}
    Macro-avg & 32 & 35 & 33 & 42 & 22 & 29 & 78 & 60 & 68 \\
    Micro-avg & 34 & 34 & 34 & 26 & 17 & 21 & 76 & 76 & 76 \\
\bottomrule
\end{tabular}
  \caption{Performance for classification from vision, audio, and
    transfer learning from text (training set of EmoInt).}
  \label{tab:perfdetails}
\end{table*}

\section{Results}
\subsection{Facial Expressions and Audio}
Table~\ref{tab:confusion} shows the confusion matrices for facial and
audio emotion recognition on our complete AMMER data set and
Table~\ref{tab:perfdetails} shows the results per class for each
method, including facial and audio data and micro and macro averages.
The classification from facial expressions yields a macro-averaged \F
score of 33\,\% across the three emotions joy, insecurity, and
annoyance (P=0.31, R=0.35). While the classification results for joy
are promising (R=43\,\%, P=57\,\%), the distinction of insecurity and
annoyance from the other classes appears to be more challenging.

Regarding the audio signal, we observe a macro \F score of 29\,\%
(P=42\,\%, R=22\,\%). There is a bias towards negative emotions, which
results in a small number of detected joy predictions
(R=4\,\%). Insecurity and annoyance are frequently confused.

\begin{table}
  \centering
  \setlength{\tabcolsep}{12pt}
  \begin{tabular}{lrrrrrrrrrrrr}
    \toprule
                 & & \multicolumn{3}{c}{Out-of-domain} \\
                 \cmidrule(l){3-5}
    Train Corpus & \ccolr{In-Domain} & \ccolr{Simple} & \ccolr{Joint C.} & \ccolr{Transfer L.}  \\
    \cmidrule(r){1-1}\cmidrule(rl){2-2}\cmidrule(rl){3-3}\cmidrule(rl){4-4}\cmidrule(rl){5-5}
    Figure8  & 66&55&59&76 \\
    EmoInt       & 62&48&56&76 \\
    TEC          & 73&55&58&76 \\
    ISEAR        & 70&35&59&72 \\
    AMMER        & 57&---&---&--- \\
    \bottomrule
  \end{tabular}
  \caption{Results in micro \F for Experiment 1 (in-domain), Experiment 2 and 3
    (out-of-domain with and without transfer learning).}
  \label{tab:exp1exp2exp3}
\end{table}

\subsection{Text from Transcribed Utterances}
The experimental setting for the evaluation of emotion recognition
from text is as follows: We evaluate the BiLSTM model in three
different experiments: (1) in-domain, (2) out-of-domain and (3)
transfer learning. For all experiments we train on the classes
\textit{anger}/\textit{annoyance}, \textit{fear}/\textit{insecurity}
and \textit{joy}.  Table~\ref{tab:exp1exp2exp3} shows all results for
the comparison of these experimental settings.

\subsubsection{Experiment 1: In-Domain application}

We first set a baseline by validating our models on established
corpora. We train the baseline model on 60\,\% of each data set listed
in Table~\ref{tab:datasets} and evaluate that model with 40\,\% of the
data from the same domain (results shown in the column ``In-Domain''
in Table~\ref{tab:exp1exp2exp3}). Excluding AMMER, we achieve an
average micro \F of 68\,\%, with best results of F$_1$=73\,\% on
TEC. The model trained on our AMMER corpus achieves an F1 score of
57\%. This is most probably due to the small size of this data set and
the class bias towards \textit{joy}, which makes up more than half of
the data set. These results are mostly in line with~\newcite{Bostan2018}.

\subsubsection{Experiment 2: Simple Out-Of-Domain application}
Now we analyze how well the models trained in Experiment 1 perform
when applied to our data set. The results are shown in column
``Simple'' in Table~\ref{tab:exp1exp2exp3}. We observe a clear drop in
performance, with an average of F$_1$=48\,\%. The best performing
model is again the one trained on TEC, en par with the one trained on
the Figure8 data. The model trained on ISEAR performs second best
in Experiment 1, it performs worst in Experiment 2.

\subsubsection{Experiment 3: Transfer Learning application}
To adapt models trained on previously existing data sets to our
particular application, the AMMER corpus, we apply transfer
learning. Here, we perform leave-one-out cross validation. As
pre-trained models we use each model from Experiment~1 and further
optimize with the training subset of each crossvalidation iteration of
AMMER. The results are shown in the column ``Transfer L.'' in
Table~\ref{tab:exp1exp2exp3}. The confusion matrix is also depicted in
Table~\ref{tab:confusion}.

With this procedure we achieve an average performance of F$_1$=75\,\%,
being better than the results from the in-domain Experiment 1. The
best performance of F$_1$=76\,\% is achieved with the model
pre-trained on each data set, except for ISEAR. All transfer learning
models clearly outperform their simple out-of-domain counterpart.

To ensure that this performance increase is not only due to the
larger data set, we compare these results to training the
model without transfer on a corpus consisting of each corpus together
with AMMER (again, in leave-one-out crossvalidation). These results
are depicted in column ``Joint C.''. Thus, both settings, ``transfer
learning'' and ``joint corpus'' have access to the same information.

The results show an increase in performance in contrast to not using
AMMER for training, however, the transfer approach based on partial
retraining the model shows a clear improvement for all models (by 7pp
for Figure8, 10pp for EmoInt, 8pp for TEC, 13pp for ISEAR)
compared to the ''Joint'' setup.

\section{Summary \& Future Work}
We described the creation of the multimodal AMMER data with emotional
speech interactions between a driver and both a virtual agent and a
co-driver. We analyzed the modalities of facial expressions,
acoustics, and transcribed utterances regarding their potential for
emotion recognition during in-car speech interactions. We applied
off-the-shelf emotion recognition tools for facial expressions and
acoustics. For transcribed text, we developed a neural network-based
classifier with transfer learning exploiting existing annotated
corpora. We find that analyzing transcribed utterances is most
promising for classification of the three emotional states of joy,
annoyance and insecurity.

Our results for facial expressions indicate that there is potential
for the classification of joy, however, the states of annoyance and
insecurity are not well recognized. Future work needs to investigate
more sophisticated approaches to map frame predictions to sequence
predictions. Furthermore, movements of the mouth region during speech
interactions might negatively influence the classification from facial
expressions. Therefore, the question remains how facial expressions
can best contribute to multimodal detection in speech interactions.

Regarding the classification from the acoustic signal, the application
of off-the-shelf classifiers without further adjustments seems to
be challenging. We find a strong bias towards negative
emotional states for our experimental setting. For instance, the
personalization of the recognition algorithm (\eg, mean and standard
deviation normalization) could help to adapt the classification for
specific speakers and thus to reduce this bias. Further, the acoustic
environment in the vehicle interior has special properties and the
recognition software might need further adaptations.

Our transfer learning-based text classifier shows considerably better
results. This is a substantial result in its own, as only one previous
method for transfer learning in emotion recognition has been proposed,
in which a sentiment/emotion specific source for labels in
pre-training has been used, to the best of our
knowledge~\cite{Felbo2017}. Other applications of transfer learning
from general language models include
\cite[\emph{i.a.}]{Rozental2018,Chronopoulou2018}.
Our approach is substantially different, not being trained on a huge
amount of noisy data, but on smaller out-of-domain sets of higher
quality. This result suggests that emotion classification systems
which work across domains can be developed with reasonable effort.

For a productive application of emotion detection in the context of
speech events we conclude that a deployed system might perform best
with a speech-to-text module followed by an analysis of the
text. Further, in this work, we did not explore an ensemble model or
the interaction of different modalities. Thus, future work should
investigate the fusion of multiple modalities in a single classifier.

\section*{Acknowledgment}
We thank Laura-Ana-Maria Bostan for discussions and data set
preparations. This research has partially been funded by the German
Research Council (DFG), project SEAT (KL 2869/1-1).

\end{document}